\documentclass[11pt]{article}

\usepackage[final]{acl}

\usepackage{times}
\usepackage{latexsym}

\usepackage[T1]{fontenc}
\usepackage[utf8]{inputenc}

\usepackage{microtype}

\usepackage{inconsolata}

\usepackage{graphicx}
\usepackage{subcaption}
\usepackage{caption}
\usepackage{stfloats} % improved placement for two-column floats
\usepackage[section]{placeins} % prevents floats from moving past a \section

\usepackage[normalem]{ulem} % sout, uline
\usepackage{xcolor}

\usepackage[disable]{todonotes}
\usepackage{booktabs}
\usepackage{multirow}
\usepackage{makecell}
\usepackage{tcolorbox}
\usepackage{listings}
\usepackage{longtable}
\usepackage{array}
\usepackage[noabbrev,capitalize]{cleveref}

\title{\textsc{ImgMT}: Wikipedia-based Dataset For Evaluating Text-in-Image Translation}

\author{Šárka Uramová, Petar Kirilov Petrov, Josef Jon, Ondřej Bojar and Michal Novák \\
         Charles University \\ Faculty of Mathematics and Physics \\ Prague, Czech Republic \\  \texttt{\{sarka.uramova458, petar.petrov242\}@student.cuni.cz} \\
         \texttt{\{jon, bojar, mnovak\}@ufal.mff.cuni.cz}
}

\begin{document}
\maketitle

\begin{abstract}
Text-in-image translation aims to translate all textual content embedded in an image. In Text Image Translation (TIT), the output is translated text; In-Image Machine Translation (IIMT) additionally renders the translation back into the image. We introduce \textsc{ImgMT}, a massively multilingual dataset and benchmark mined from localized SVG illustrations on Wikimedia Commons, focusing on diagrams, maps, and infographics. \textsc{ImgMT} contains 35,612 aligned image pairs grouped into 1,855 image sets, spanning 129 languages, 20+ writing systems, and 4,886 language pairs.

We propose a multi-level evaluation protocol that measures source-side text-region detection and OCR accuracy alongside target-side translation quality. We benchmark several multimodal large language models and a dedicated TIT system in English$\rightarrow$X and X$\rightarrow$English settings, and find that the source language and writing system substantially affect TIT quality. Finally, we show that \textsc{ImgMT} is not well suited for isolating the contribution of visual information in multimodal translation.
\end{abstract}

\section{Introduction}
\label{sec:intro}

Modern multimodal models can translate from text, images, or both.
A family of \emph{text-in-image translation} tasks takes an image as input and aims to translate all textual elements appearing within it.
This setting includes Text Image Translation (TIT), which outputs translated text (optionally with bounding boxes), and In-Image Machine Translation (IIMT), which further renders the translated text back into the image.
By contrast, in Multimodal Machine Translation (MMT), the source is text and the image is used only as auxiliary context.

Several datasets for text-in-image translation already exist \citep{ma.etal22,lan.etal23,qian.etal24,lyu.etal25,vaidya.etal25,wu.etal26}, but they share a key limitation: limited language coverage. Even the most multilingual resources typically include only 14--16 languages \citep{li.etal25,Li_2026_CVPR}. Consequently, existing benchmarks do not sufficiently stress-test systems' multilingual capabilities, either in translating across a wide range of languages or in recognizing text written in diverse scripts.

Wikipedia has served as a rich source of multilingual supervision for multimodal tasks \citep[e.g.,][]{srinivasan.etal21}. For vision-language tasks, a particularly appealing property is that many Wikipedia illustrations are localized: the same visual is available with labels translated into dozens of languages. Since these illustrations are often stored as SVGs, where text is encoded separately from the graphical elements, they can be mined to create aligned multilingual resources for text-in-image translation.

\begin{figure}[t]  %TODO fix
    \centering
%    \begin{subfigure}[b]{0.49\textwidth}
        \centering
        \includegraphics[width=\linewidth]{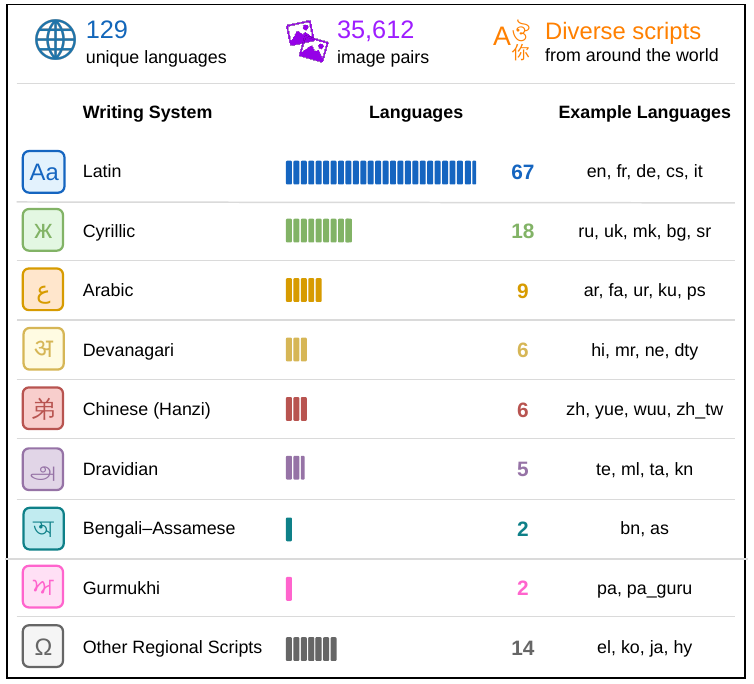}
       % \caption{}
    \caption{\textsc{ImgMT} languages across 
different scripts.}
        \label{fig:lang_distribution}
%    \label{fig:intro_images}
\end{figure}

In this paper, we introduce \textsc{ImgMT}, a dataset and evaluation framework for assessing methods for text-in-image translation. \textsc{ImgMT} contains 35,612 aligned image pairs grouped into 1,855 sets of visually distinct SVG images extracted from Wikimedia Commons. The embedded text spans 129 languages, covering more than 20 writing systems (as illustrated in \cref{fig:lang_distribution}), and yields 4,886 distinct language pairs.

\begin{figure}[t]  %TODO fix
%    \end{subfigure}
%    \hfill
%    \begin{subfigure}[b]{0.49\textwidth}
        \centering
        \includegraphics[width=\linewidth]{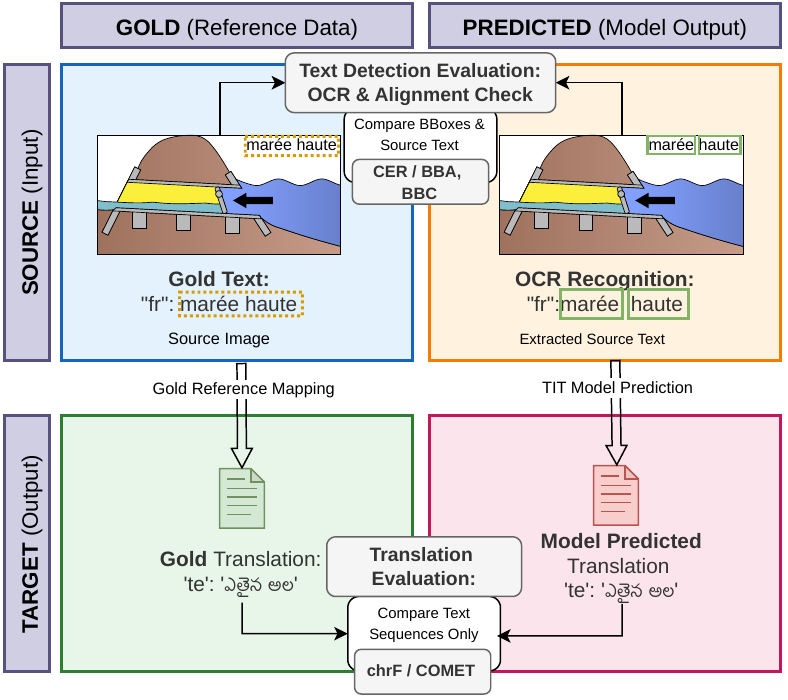}
        \caption{Text Image Translation evaluation diagram.}
        \label{fig:tit_evaluation}
%    \end{subfigure}
%    \caption{\textsc{ImgMT} evaluation protocol.}
%    \label{fig:intro_images}
\end{figure}

\textsc{ImgMT} targets diagrams, maps, and infographics---content typically created as vector graphics. In contrast to scene text translation \citep{lan.etal23,qian.etal24,vaidya.etal25}, where robustness to distorted or occluded text is often the primary challenge, our setting is dominated by translation-specific difficulties.
Unlike document image translation \citep{liang.etal24}, where translated passages can be substantially longer, the text snippets in \textsc{ImgMT} are often short and information-dense, and they frequently include specialized terminology and named entities from scientific and educational domains.

Along with the dataset, we release an evaluation framework.
As shown in \cref{fig:tit_evaluation}, it evaluates system outputs on both the source and target sides: on the source side, it measures text-region detection and OCR accuracy; on the target side, it scores translations using standard MT metrics.

To demonstrate our evaluation framework in practice, we use it to benchmark multiple TIT systems, most of which are general-purpose vision--language models (VLMs).\footnote{The framework also supports IIMT evaluation, but we leave it for future work.}
We evaluate their ability to translate from English into a wide range of target languages.
In addition, we benchmark one system on translation into English to study how different source-side writing systems affect performance.
%\formn{AnyTrans experiments?}

Beyond evaluating TIT systems, we ask whether \textsc{ImgMT} is also useful for evaluating multimodal machine translation, specifically for quantifying the contribution of the visual modality to translation quality. In many infographics, the extracted text is sufficiently self-contained to translate correctly on its own, especially when the system is given all text at once. Therefore, any improvement from providing the image may reflect either genuine visual grounding (e.g.\ resolving polysemy, units, or abbreviations using surrounding graphics) or incidental cues such as layout and line segmentation. To tease these factors apart, we compare translations produced with (i) the full source image, (ii) a text-only rendering that preserves layout while removing graphical content, and (iii) plain text input without any image.

The \textsc{ImgMT} dataset and evaluation framework are publicly available.\footnote{Released simultaneously on Hugging Face (\url{https://huggingface.co/datasets/sarkau/ImgMT-Dataset-For-Evaluating-Text-Image-Translation}) and in the LINDAT/CLARIAH-CZ repository (\url{http://hdl.handle.net/11234/1-6251}).}
We also provide a development repository\footnote{\url{https://github.com/ufal/imgmt-devel}} with supplementary scripts and intermediate artifacts produced during this work.

The remainder of the paper is organized as follows.
We first situate \textsc{ImgMT} in the context of related work (\cref{sec:relwork}). We then introduce the dataset (\cref{sec:data}) and describe our evaluation protocol (\cref{sec:evaluation}). Next, we detail our experimental setup and how we use \textsc{ImgMT} to benchmark TIT approaches (\cref{sec:expersetup}), followed by the benchmarking results (\cref{sec:results}).
Finally, we conclude in \cref{sec:concl}.

\section{Related Work}
\label{sec:relwork}

\subsection{Datasets for text-in-image translation}
Existing datasets for TIT and IIMT vary widely in realism (synthetic vs.\ authentic), supervision (text-only vs.\ image-to-image), and domain.
Synthetic resources such as E2E\_TIT\_With\_MT \citep{ma.etal22}, the Multilingual IIMT dataset \citep{lan.etal24}, IIMT30k \citep{tian.etal25}, and VT-Syn \citep{vaidya.etal25} scale by rendering parallel text onto background images, producing large corpora with controllable variation; their synthetic nature makes it straightforward to provide both source- and target-language texts and images.

Authentic datasets instead curate real images with natural typography and visual clutter, typically relying on OCR and MT during construction. They often omit target-language images, which limits their direct applicability to IIMT and prevents evaluation in the reverse translation direction.
OCRMT30K \citep{lan.etal23}, MTIT6 \citep{qian.etal24}, and VT-Real \citep{vaidya.etal25} collect human-translated scene text from multiple OCR datasets, with OCRMT30K and MTIT6 additionally providing bounding boxes. Although MTIT6 does not include target images, it supports IIMT evaluation by applying OCR to system-generated target images to recover the translated text.

ECOIT \citep{zhu.etal23} and MIT-10M \citep{li.etal25} target e-commerce imagery crawled from the web, pairing OCR-extracted text with MT; ECOIT uses human post-editing, while MIT-10M applies a secondary MT system and embedding-based semantic similarity to filter likely translation errors.
Following a similar pipeline, IMTBench/IMT-1M \citep{lyu.etal25} and MMTIT-Bench \citep{Li_2026_CVPR} combine scene images, web-crawled images, and document images.
DoTA \citep{liang.etal24} focuses on document-page translation, using images of scientific-paper pages and providing outputs in Markdown.

Vectra-Bench \citep{wu.etal26} is one of the few benchmarks with authentic source images that also provides target-language images (generated by vision--language models), enabling direct IIMT evaluation; it further emphasizes evaluation beyond text accuracy, including multilingual generalization and visual quality.

Most existing datasets cover only a small set of languages, with the broadest coverage (14--16 languages) reported by \citet{li.etal25,Li_2026_CVPR}. In addition, because target-side images are often missing, these datasets are generally unsuitable for reverse-direction translation.

\textsc{ImgMT} differs from prior datasets along three axes. First, it targets vector-graphic diagrams, maps, and infographics rather than natural scene photos, e-commerce images, or document pages, shifting the challenge toward translating short, information-dense labels with specialized terminology and named entities. Second, it is built from Wikipedia SVG variants that preserve the underlying visual content while changing only the text, yielding aligned multilingual image pairs with consistent layouts and text positions, and directly supporting IIMT. Third, it substantially expands multilingual breadth: \textsc{ImgMT} spans 129 languages and over 20 writing systems, enabling evaluation across 4,886 language pairs in both directions and stress-testing multilingual and multi-script text-in-image translation beyond the coverage of prior benchmarks.

%\subsection{Approaches to TIT}
%\forpp{if you have any overview from the NPFL101 presentation, put som ideas here}
%\formn{extract the overview from the other papers}

\subsection{Effect of Image in Multimodal Translation}
\label{sec:effect_of_image}

The question of whether visual context improves translation quality has been studied extensively in the general MMT literature, most prominently in the context of the Multi30K dataset \citep{elliott-etal-2016-multi30k} and the associated WMT Multimodal Translation shared tasks \citep{parida-etal-2024-findings}. CoMMuTE \citep{futeral.etal23} is an evaluation dataset for resolving textual ambiguity with visual context: 155 ambiguous English sentences, each paired with two images and two meaning-specific translations into French, German, and Czech by native speakers, with images sourced from Google Images or the authors.

Early architectures such as multi-encoder, double-attention or gated fusion models reported only marginal improvements from incorporating image features over strong text-only baselines, and subsequent adversarial evaluations \citep{elliott-2018-adversarial} showed that many multimodal systems are largely insensitive to the visual input, achieving similar scores even when paired with unrelated or shuffled images. \citet{caglayan-etal-2019-probing} revisited this question and found that visual grounding becomes useful mainly when the textual signal is degraded or ambiguous, rather than in standard, fully legible text conditions. More recently, \citet{wu-etal-2021-good} argued that much of the previously reported benefit of visual context in prior work stemmed from erroneous evaluation protocols rather than from genuine multimodal grounding.

Recent work with large vision-language models suggests that visual context is useful mainly when it resolves information that is missing or ambiguous in the text. DeDiT \citep{liu-etal-2025-detect}, for example, first detects textual ambiguity and then, optionally, uses the visual modality. Similarly, \citet{yu-etal-2025-imagination} argue for adaptive fusion because images can help with ambiguity but may also introduce redundant or noisy information. Most recently, \citet{guerra-khrapunova-2026-picture} show that the benefits of visual context are strongly task-dependent: relevant images help on CoMMuTE, but gains do not transfer consistently to other domains, including technical documentation with accompanying figures.

This distinction is especially relevant for diagrams, maps, and infographics. Visual structure may be important for locating text, recovering reading order, or linking labels to graphical elements, but once the text is correctly extracted and semantically self-contained, the surrounding image may add little to the translation itself. We therefore expect visual context to matter mainly for ambiguous labels, abbreviations, or domain-specific terms whose meaning depends on the graphic, while having little effect on ordinary text. \Cref{sec:no_visual_modality} tests this hypothesis directly.

\section{Dataset}
\label{sec:data}

The primary source of data for this project is Wikimedia Commons,\footnote{\url{https://commons.wikimedia.org/}} a repository of freely usable images and other multimedia content related to~Wikipedia topics. Our objective was to create a multilingual dataset consisting of image pairs that differ only in their textual content, where texts at~identical positions represent mutual translations.

%\forsu{Brief structural description}
At the top level, the dataset is split into the directories \texttt{train}, \texttt{dev} and \texttt{test}. Within each split, files are organized into numbered folders representing image groups. Each group directory corresponds to~one Wikimedia image with all its variants.\footnote{We prefer to talk about image \emph{groups} rather than simply images because the image files belonging to the same group can and do differ. Conceptually, they should be the same image, differing at least in the text translation variant.} It contains SVG source files, their PNG exports, and alignment annotations stored in JSON format. Each JSON file represents a directed translation pair between a source and a target language,\footnote{It can also be used for translation in the reverse direction.} specifying image metadata, bounding box coordinates and aligned texts. For~detailed overview of the repository directory and JSON schema, see Appendix~\ref{sec:app-dataset-structure}.

\subsection{Basic Statistics}
The dataset contains 35,612 aligned image pairs across 1,855 image groups (i.e., visually distinct images), comprising 235,637 text strings in total. The number of language variants per group ranges from 2 to 77, with an average of~4.95~variants per group.

Overall the dataset includes 129 unique languages in over 20 writing systems (such as Latin, Cyrillic, Arabic, Devanagari, etc.; see more in~\cref{fig:lang_distribution}). Of 8,256 possible language pairs,\footnote{Corresponding to (${129\choose 2}$).} the dataset includes 4,886 that can be used in both translation directions.

\paragraph{Theoretical vs. actual image pairs}
During the~initial crawling (see \cref{sec:pipeline}), we collected raw URLs for all available image language variants in each group. Theoretically, taking any language as the source and any language as the target, this could lead up to 118k image pairs.
However, web-scraped SVG files are often noisy or unsuitable for our purpose. After applying our pipeline, the final dataset contains 35k pairs (about 30\% of~the theoretical maximum). The~drop is intentional: many pairs could not be created because of~issues in the SVG files, such as (1) invalid SVGs or SVGs with zero height or width, (2) font rendering errors, (3) failing the spatial overlap check, and (4) text saved as~paths (\texttt{<path>}) instead of text elements (\texttt{<text>}) – unlike normal text elements, outlined paths do not store characters or words; they are saved only as~drawing curves (shapes). 
While we could use OCR to read these shapes, we decided not to do so to avoid introducing recognition errors into the gold ground-truth data.

%\forsu{Languages. The most populated (how many pairs), the least populated (how many pairs).}
\paragraph{Language distribution}
The dataset contains a~wide variety of languages, but they are not evenly distributed. The most common languages are English (appearing in 5k image pairs across the entire dataset), Spanish (with 3.5k image pairs) and French (appearing in~3.3k pairs). In contrast, some languages are very rare, for example, Northern Sami language and Venetian language appear in only 4 pairs within the~whole dataset. The complete distribution is detailed in Appendix~\ref{app:langs}.

%\forsu{English. How many pairs contain English. Especially needed for the test set}
The test split is similar: 3,575 image pairs in total and 548 of them (15\%) involve English. This allows us to reliably test translations to and from English while still keeping other diverse languages in the~test set.

\begin{table}[t]
\centering
\small
\begin{tabular}{lrr}
\toprule
\textbf{Category} & \textbf{Count} & \textbf{Share}\\
\midrule
Proper name & 126,835 & 60.2\%\\
Common noun phrase & 79,175 & 37.6\%\\
Statement without finite verb & 1,964 & 0.9\%\\
List of common noun phrases & 1,003 & 0.5\%\\
Single clause & 943 & 0.4\%\\
List of proper names & 493 & 0.2\%\\
List mixing common and proper & 127 & 0.1\%\\
Multiple clauses & 78 & 0.0\%\\
List of statements without finite verb & 26 & 0.0\%\\
\midrule
\textbf{Consensus total} & \textbf{210,644} & \textbf{100.0\%}\\
No consensus (disagreed labels) & 24,993 & --\\
\textbf{All strings} & \textbf{235,637} & --\\
\bottomrule
\end{tabular}
\caption{Distribution of string types (LLM-based categorization). Shares are computed over the consensus subset.}
\label{tab:string-types}
\end{table}

\paragraph{String type distribution.}
Because \textsc{ImgMT} consists primarily of diagrams, maps, and infographics, we expect most extracted strings to be short and rarely to form full sentences. Across all 235,637 strings, the average length is 9.29 characters.
To characterize the strings more deeply, we performed a multilingual analysis using an ensemble of three LLMs (Gemma 4, GPT-OSS-120B, Qwen3.8-27b). The models agreed on a label for 210,644 strings (89.4\%; see \cref{tab:string-types}), while the remaining 24,993 strings had no unanimous label.
Across the full set of strings, about 60\% are proper names, about 33--38\% are common noun phrases, and 1--3\% are statements without a finite verb; well-formed sentences account for only about 0.5\%.
Overall, this distribution indicates that \textsc{ImgMT} is dominated by short, label-like text, with a high share of proper names rather than sentence-level discourse.

\subsection{Processing Pipeline} \label{sec:pipeline}
\begin{figure*}[t]
    \centering
    \includegraphics[width=\textwidth]{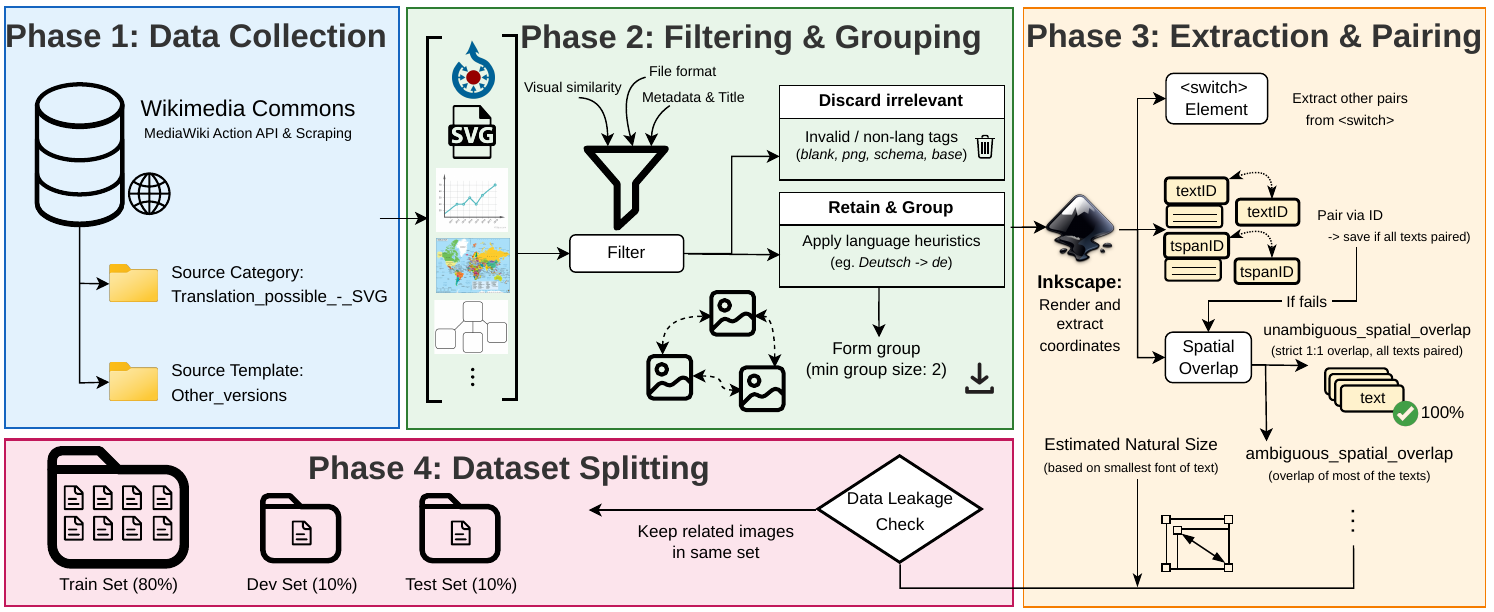}
    \caption{Pipeline overview for generating the multilingual \textsc{ImgMT} dataset}
    \label{fig01:processing_pipeline}
\end{figure*}

As shown in \cref{fig01:processing_pipeline}, the~pipeline consists of~the~four following steps: (1) data collection, (2)~filtering and grouping, (3) extraction and pairing and (4) dataset splitting.

\paragraph{Data collection}
In this stage, raw multimodal data is sourced from Wikimedia Commons. We primarily crawl files from the category \textit{Translation\_possible\_-\_SVG},\footnote{\url{https://commons.wikimedia.org/wiki/Category:Translation_possible_-_SVG}} which is a rich source of multilingual diagrams, maps, graphs etc. Additionally, we extract explicitly linked language variants by parsing the \textit{Other versions} templates often found on file description pages. This combined approach allows us to get a large initial pool of potential image translations. 

\paragraph{Filtering and grouping}
The crawled data goes through a filtering process. Non-SVG formats or SVG files which cannot be parsed and images that render with zero width or height are immediately discarded. Then we parse the file descriptions to~extract valid language codes using a custom heuristic mapped to~ISO language codes. While images from the \textit{Translation\_possible\_-\_SVG} category often label their variants by language, this is not always the case (some labels describe file types or visual changes rather than languages).
Next, to accurately group the translated variants, we evaluate visual similarity across the images. Candidate files found within the same translation table on the image Wikimedia page are grouped together only if their visual layout is similar, evaluated using the \emph{Structural Similarity Index Measure (SSIM)} \citep{wang2004ssim}.
%\forsu{what exactly do you use for checking vis sim?}
If a visually consistent cluster contains two or more valid language variants, we keep it as a separate group.

\paragraph{Extraction and pairing}
To accurately extract textual elements and their spatial coordinates, we utilize Inkscape \citep{inkscape} which renders the SVGs to provide precise absolute bounding boxes. The extracted text elements are then paired across languages using three prioritized strategies. 
First, if the SVG file contains \texttt{<switch>} elements, texts are paired purely based on predefined \texttt{systemLanguage} tags (the \texttt{svg\_switch} method). 
Second, if standard switches are absent, the pipeline attempts to align texts using structural identifiers (eg. \texttt{tspanID} or \texttt{textID}; the \texttt{svg\_id\_match} strategy).
However, as internal IDs are frequently mismatched, missing or incomplete across translated images, we mostly rely on the third method: spatial overlap, which pairs text elements based on~their intersection over union (IoU score). The~image pairs are then categorized as \emph{unambiguous} (strict 1:1 match between the boxes with zero unmapped ones; denoted \texttt{unamb\_spatial\_overlap}) and \emph{ambiguous} mapping (the rest; labeled \texttt{amb\_spatial\_overlap}).
\cref{tab:dataset_statistics} reports detailed statistics on the number of image pairs obtained by each alignment method in the cascade.
Finally, the image dimensions and corresponding bounding boxes are normalized: we identify the minimum font size within an image group and proportionally scale the image dimensions and spatial coordinates by~a~calculated ratio, ensuring that the smallest text remains clearly legible.

\begin{table}[t]
\centering
\small
\begin{tabular}{l r r r}
\toprule
\textbf{Alignment Method} & \textbf{Train} & \textbf{Dev} & \textbf{Test} \\
\midrule
1. \texttt{svg\_switch} & 6,514 & 238 & 328 \\
2. \texttt{svg\_id\_match} & 366 & 41 & 28 \\
3. \texttt{unamb\_spatial\_overlap} & 8,842 & 1,223 & 1,065 \\
4. \texttt{amb\_spatial\_overlap} & 12,771 & 2,042 & 2,154 \\
\midrule
\multicolumn{3}{l}{\textit{Theoretically achievable image pairs}} & 118,000 \\
\multicolumn{3}{l}{Total JSON metadata files} & 35,612 \\
\multicolumn{3}{l}{Total PNG images} & 9,191 \\
\bottomrule
\end{tabular}
\caption{Success of text alignment methods across dataset splits. The methods are carried out in the specified order and an image pair is counted with the method which succeeds first.}
\label{tab:dataset_statistics}
\end{table}

\paragraph{Dataset splitting}
In the final stage, the processed data is distributed into \texttt{train}, \texttt{dev} and \texttt{test} sets using an 80:10:10 ratio.\footnote{Due to the leakage-prevention clustering constraint, the resulting dev and test splits each comprise slightly more than 10\% of the total data.} Importantly, this stage implements a data leakage prevention mechanism. Image groups that share identical visual content but differ slightly in layout or metadata — such as~a~standard historical map versus a visually impaired version, which might have been initially parsed as~a~separate groups — are identified and put into the~same dataset split. This constraint guarantees that no language variant or minor visual modification of~a~specific image seen during training can appear in the~development or test sets. The~splitting algorithm also balances the distribution of the three alignment methods across all sets, as detailed in \cref{tab:dataset_statistics}. However, to strictly prevent cross-split visual leakage, we did not enforce explicit balancing of individual languages or writing systems across the sets.

\paragraph{Producing the PNG images}
To convert the processed SVG files into raster PNG images, we use Inkscape via command-line interface. To maintain identical resolution and scale across all language variants within a~single group, the target export height is extracted from the alignment JSON metadata (which was pre-calculated based on the smallest font size, as described above) and passed to~Inkscape using the \texttt{--export-height} flag.

%Our rendering pipeline also handles multi-language single-file SVGs using the \texttt{<switch>} tag. For these files, the script parses the document, matches each \texttt{systemLanguage} attribute to isolate individual languages, removes the attribute from the selected node so Inkscape renders it properly and generates a separate PNG image for every supported language. 

\subsection{Manual Evaluation of the Dataset Collection Method}
\label{sec:maneval}

The dataset construction pipeline comprises multiple stages, each of which can introduce errors. For example, paired images can diverge beyond acceptable text differences, some text may be missed when it is encoded as vector paths rather than text objects, and our position-based heuristics can fail. To quantify the impact of these issues, we compare the automatically collected dataset against manual annotations.

%\forcamera{Those 10 missing images were omitted because of a bug in the tool related to processing switch SVGs. I expect the numbers to increase once we fix it. Fix the bug and annotate}
%\forcamera{Annotate all image pairs annotated as different}

We randomly sample 100 image pairs from the test set and annotate them using a purpose-built tool implemented for this study (a screenshot is shown in Appendix~\ref{app:annottool}).%
\footnote{%
Due to a display bug affecting SVGs paired using the \texttt{svg\_switch} method, we had to omit 10 such examples from the evaluation. Since this alignment method is, by design, the most precise, this omission should not negatively affect the reported statistics.
}
First, annotators judge whether the images in the pair are identical (still regardless the texts). They may decide to skip an example if they consider the images too different. Second, annotators manually correct the bounding boxes and their cross-image correspondences. All bounding-box edits are permitted (addition, deletion, and resizing). Annotators do not edit the text content associated with each box, as this is often impractical, particularly when the text uses an unfamiliar script. Finally, the annotators correct the cross-lingual correspondences. These are required to be one-to-one, which may require splitting or merging bounding boxes. In addition to visual cues, annotators can use automatic English translations of the extracted text fields provided by the annotation tool. Two annotators split the batch evenly, each correcting half of the pairs.

In 13\% cases the paired images were too different for text-in-image translation evaluation.
In 16\% cases the images were identical except for minor graphical differences. These images are usable for IIMT with a grain of salt but they are perfectly suitable for TIT evaluation.
The remaining pairs of images were visually completely identical.

We measure bounding-box localization accuracy with two precision/recall metrics that quantify alignment between the manually annotated boxes and those in the original data.
\emph{Bounding Box Count} (BBC) measures agreement in the number of boxes, while \emph{Bounding Box Area} (BBA) measures agreement by the area of overlap between boxes.

\cref{tab:manual-eval} reports bounding-box quality for all boxes (text-field detection), matched boxes (a one-to-one match exists across the pair), and correctly matched boxes (the match is the correct cross-lingual correspondence).

After all steps, BBA precision remains above 91\%, while recall is around 83\%; most of the gap is due to text-field detection.
BBC exhibits similar performance.
Although this recall drop should not affect comparing multiple systems on the same dataset, it should be considered when interpreting the absolute results.

\begin{table}[t]
  \centering
  \resizebox{\columnwidth}{!}{
  \begin{tabular}{lcccccc}
    \toprule
    \multirow{2}{*}{\makecell[l]{\textbf{Bounding}\\\textbf{boxes}}} & \multicolumn{3}{c}{\textbf{Count}} & \multicolumn{3}{c}{\textbf{Area}} \\
    \cmidrule(lr){2-4} \cmidrule(lr){5-7}
    & \textbf{P} & \textbf{R} & \textbf{F1} & \textbf{P} & \textbf{R} & \textbf{F1} \\
    \midrule
    All boxes & 96.2\% & 90.6\% & 91.5\% & 94.9\% & 86.8\% & 89.0\% \\
    Matched boxes & 96.2\% & 90.2\% & 91.3\% & 94.9\% & 86.6\% & 88.9\% \\
    Correctly matched & 92.7\% & 86.9\% & 88.3\% & 91.6\% & 83.3\% & 86.0\% \\
    \bottomrule
  \end{tabular}
  }
  \caption{Manual evaluation of dataset quality. Rows correspond to (i) all detected text-field boxes (\emph{All boxes}), (ii) boxes for which a one-to-one cross-image correspondence exists (\emph{Matched boxes}), and (iii) boxes whose correspondence is the correct cross-lingual match (\emph{Correctly matched}). Columns report precision/recall/F1 for box \emph{count} (BBC) and \emph{area overlap} (BBA).}
  \label{tab:manual-eval}
\end{table}

\section{Evaluation Method}
\label{sec:evaluation}

In our work, we focus on evaluating Text-in-Image Translation (TIT). As shown in Figure~\ref{fig:tit_evaluation}, our evaluation system compares gold reference data with model predictions (machine output).

\paragraph{Text detection and OCR evaluation}
The first phase evaluates the model's ability to accurately detect and read text from the source image. We compare the extracted predicted text with the source gold text using the \emph{Character Error Rate (CER)}.

If predicted bounding boxes are available – which may not be the case for models that only output raw text – we use them to pair the predicted texts with the reference texts. At the same time, we evaluate the bounding boxes localization accuracy by two metrics already introduced in \cref{sec:maneval}: (1)~\emph{BBA} (\textit{Bounding Box Area}), which measures spatial overlap by calculating the harmonic mean of area-based precision and recall across intersecting bounding boxes, and (2)~\emph{BBC} (\textit{Bounding Box Count}), which measures discrete block-level retrieval based on paired True Positives.
If not stated otherwise, we mean the F1 score by referring to these measures. 

To handle cases where an OCR model splits a~single line or paragraph into multiple smaller boxes, we use a text merging mechanism. If several predicted boxes overlap with the same reference box, we sort their text horizontally from left to right and join it into one string. We also merge their coordinates into a single bounding box before calculating the CER and BBA metrics.

\paragraph{Translation evaluation}
The second phase measures the quality of translated text. We evaluate the output using a traditional lexical metric \emph{chrF} \citep{popovic2015chrf} and an advanced semantic metric \emph{COMET}, specifically the \texttt{wmt22-comet-da} model \citep{rei-etal-2022-comet}.

\paragraph{Text-based alignment}
In scenarios where the~prediction does not provide bounding box coordinates, the system aligns text blocks using the Hungarian algorithm for matching in bipartite graphs. Alignment is driven by a composite cost matrix combining text dissimilarity with a spatial reading-order penalty. Dissimilarity is calculated using CER for OCR or inverted quality score (one minus the metric score) for translation evaluation via chrF or COMET, with low-quality candidate pairs filtered out by task-specific thresholds. To maintain spatial consistency, we add a position penalty calculated as the absolute difference between the~relative positions of predicted and reference texts, assuming a top-to-bottom, left-to-right reading order. The~algorithm then selects one-to-one pairs that minimize the total cost and rejects any pairs that exceed the~threshold.

%The cost matrix combines text dissimilarity (CER for OCR, or \textit{1.0~-~score} for chrF and COMET) with a relative position penalty assuming a top-to-bottom, left-to-right reading order. Pairs that do not meet the similarity threshold are excluded from the final alignment.

\section{Experimental Setup}
\label{sec:expersetup}

To demonstrate the practical utility of the proposed framework, we evaluate several multimodal systems on both the TIT and MMT tasks. All experiments are conducted on the dataset's test split.
For TIT, we use the PNG versions of the images at their native resolution as the input, without any additional preprocessing or visual transformations.

\paragraph{Evaluation of TIT systems}
The first experiment evaluates the performance of several multimodal large language models (MLLMs) and one dedicated pipelined system on the TIT task with English as the source language.

The evaluated MLLMs are \textbf{Qwen3.5} (MoE of 122B parameters, 10B active), \textbf{Claude Haiku 4.5}, \textbf{Kimi K3} (MoE of 2.8T parameters, 104B active), and \textbf{GPT-5.6 Luna}. To ensure a fair comparison, all models are evaluated using the same zero-shot prompt. The prompt instructs the model to detect all text present in the bitmap image, estimate the corresponding source text bounding boxes, transcribe the detected text, and translate it into the target language. The expected output consists of a structured representation containing the predicted source text bounding boxes, the recognized source texts, and the translated target texts.

We also evaluate AnyTrans \citep{qian.etal24}, a pipelined IIMT system that we slightly adapt to our benchmark. While \textbf{AnyTrans} can render translations back into the image, we do not use this capability in our TIT experiments. Given an input image and the source and target languages, the system performs translation in three stages. First, an OCR engine detects and recognizes the text in the image, producing transcriptions together with their bounding boxes. Our adaptation supports multiple OCR backends; in the experiments reported here, we use \emph{PaddleOCR} \citep{paddleocr} only. The recognized text is then passed to a large language model for translation; we use \emph{DeepSeek-V4-Flash-Thinking}. Finally, in the rendering stage (not used in our TIT setup), the original text regions are removed and replaced with the translated text using the detected bounding boxes.

In these experiments, an image is scored only when the model returns a well-formed output that matches the required structured format. If the model fails to produce a valid result, that image is skipped and excluded from all metrics. Reported scores therefore reflect performance on the subset of English-source test images with valid model outputs, not on the full test set.

\paragraph{Effect of source script on TIT performance}
The second experiment aims to test the influence of writing systems on TIT performance. It evaluates \textbf{GPT-5.6 Luna} on language pairs with English as the target language, allowing for comparison of translation quality across different source scripts. GPT-5.6 Luna is selected for this experiment because it combines relatively strong overall performance with a low rate of invalid outputs as demonstrated in the previous experiment, resulting in a more complete and reliable evaluation across the diverse set of source languages. Results are reported separately for each script group using the same OCR and translation metrics.

\paragraph{Effects of visual modality on translation quality}
To isolate the contribution of visual context in our MMT analysis (beyond translating the extracted text alone) we compare three input conditions on the English-source test set.
Because this experiment targets MMT, we evaluate only the four MLLMs and exclude AnyTrans.
The extracted source text---pulled directly from the SVGs (i.e., without OCR)---is held constant across conditions: (1) \emph{Image}, where the fully rendered source image is provided alongside the text input; (2) \emph{Text-only image}, where we render the same image after removing all non-text graphical elements (e.g., backgrounds, icons, shapes) and retaining only the black text layer, preserving line breaks and spatial layout but discarding graphical content (see \cref{sec:pipeline}); and (3) \emph{No image}, where the model receives only the extracted source text segments with no visual input.

\section{Experiments and Results}
\label{sec:results}

%\forpp{extend the from-EN table with AnyTrans results (1 OCR > 1 OCR + oracle OCR > 2 OCRs + oracle OCR) if it's on the same test set, otherwise use another table and describe in the text, don't forget on captions}

\paragraph{TIT from English}
Table~\ref{tab:end-to-end TIT} reports end-to-end TIT scores for all evaluated systems, covering translation quality, OCR accuracy, and bounding-box localization.
Kimi K3 performs best overall, achieving the highest chrF (61.3) and COMET (0.834). GPT-5.6 Luna attains slightly higher localization scores (BBA and BBC) while delivering comparable translation quality. Claude Haiku 4.5 lags behind on all metrics, with a CER of 0.822 and markedly lower translation and localization scores.

Qwen3.5 shows worse OCR quality and text localization, but almost the same translation quality. This suggests that end-to-end translation quality is not determined solely by exact source-side transcription and bounding-box prediction. Nevertheless, the large performance gap of Claude Haiku 4.5 across both source- and target-side metrics indicates that errors in text detection and recognition can become a major bottleneck when sufficiently severe. This is further evidenced by results on the gold source texts extracted from the SVGs in \Cref{tab:chrf-scores-no-ci} -- Claude has a similar translation performance to the other three models in that setting.

The AnyTrans pipeline achieves the lowest CER (0.161) and competitive localization. Its translation quality is slightly lower than Kimi K3, Qwen3.5, and GPT-5.6 Luna, though it remains well above Claude Haiku 4.5.

\begin{table}
    \centering
    \resizebox{\columnwidth}{!}{
    \begin{tabular}{lccccc}
    \toprule
        \textbf{Model} & \textbf{chrF} & \textbf{COMET} & \textbf{CER} & \textbf{BBA} & \textbf{BBC} \\
    \midrule
         Qwen3.5 & 59.86 & 0.827 & 0.313 & 0.453 & 0.746 \\
         Kimi K3 & 61.30 & 0.834 & 0.174 & 0.620 & 0.829 \\
         GPT-5.6 Luna & 59.72 & 0.826 & 0.224 & 0.655 & 0.838 \\
         Claude Haiku 4.5 & 39.78 & 0.681 & 0.822 & 0.203 & 0.542 \\
         AnyTrans & 57.50 & 0.803 & 0.161 & 0.620 & 0.813 \\
    \bottomrule
    \end{tabular}
    }
    \caption{Translation and OCR performance of multimodal LLMs on English-source language pairs.}
    \label{tab:end-to-end TIT}
\end{table}

\begin{table}[tbp]
\centering
% Green = improvement, Red = degradation (note: for CER, lower is better)
\newcommand{\good}[1]{\textcolor{green!50!black}{#1}}
\newcommand{\bad}[1]{\textcolor{red!70!black}{#1}}
\resizebox{\columnwidth}{!}{
\begin{tabular}{l c ccc cc}
\toprule
\textbf{Source Script} & \textbf{$N$} & \textbf{CER} ($\Delta$) & \textbf{BBA} ($\Delta$) & \textbf{BBC} ($\Delta$) & \textbf{chrF} ($\Delta$) & \textbf{COMET} ($\Delta$) \\
\midrule
Latin            & 255 & \good{-0.106} & \good{+0.028} & \good{+0.011} & \good{+7.38}  & \good{+0.035} \\
Dravidian        & 49  & \bad{+0.359}  & \bad{-0.076}  & \good{+0.005} & \bad{-26.07} & \bad{-0.129} \\
Cyrillic         & 48  & \bad{+0.094}  & \good{+0.015} & \good{+0.002} & \good{+8.90}  & \good{+0.073} \\
Devanagari       & 25  & \good{-0.002} & \bad{-0.006}  & \good{+0.027} & \bad{-3.12}  & \bad{-0.001} \\
Bengali-Assamese & 22  & \good{-0.025} & \bad{-0.035}  & \bad{-0.068}  & \bad{-15.80} & \bad{-0.098} \\
Arabic           & 13  & \good{-0.129} & \bad{-0.253}  & \bad{-0.229}  & \bad{-21.78} & \bad{-0.132} \\
Gurmukhi         & 11  & \bad{+0.178}  & \good{+0.032} & \good{+0.084} & \bad{-16.21} & \bad{-0.110} \\
Gujarati         & 9   & \bad{+0.174}  & \bad{-0.024}  & \good{+0.062} & \bad{-10.04} & \bad{-0.070} \\
Chinese          & 7   & \good{-0.003} & \good{+0.052} & \good{+0.075} & \good{+9.44}  & \good{+0.032} \\
Ol Chiki         & 5   & \bad{+1.001}  & \good{+0.055} & \good{+0.153} & \bad{-49.50} & \bad{-0.225} \\
Georgian         & 5   & \bad{+0.079}  & \good{+0.187} & \good{+0.140} & \good{+9.68}  & \good{+0.087} \\
Japanese         & 3   & \good{-0.255} & \good{+0.099} & \bad{-0.033}  & \good{+19.37} & \good{+0.104} \\
Korean           & 3   & \good{-0.232} & \bad{-0.098}  & \bad{-0.012}  & \good{+31.58} & \good{+0.197} \\
Hebrew           & 3   & \good{-0.124} & \bad{-0.205}  & \bad{-0.298}  & \good{+21.12} & \good{+0.156} \\
Burmese          & 2   & -            & \bad{-0.590}  & \bad{-0.774}  & \bad{-57.71} & \bad{-0.754} \\
Thai             & 1   & \bad{+0.165}  & \good{+0.144} & \good{+0.226} & \bad{-0.36}  & \bad{-0.028} \\
Armenian         & 1   & \bad{+0.154}  & \bad{-0.050}  & \bad{-0.679}  & \bad{-24.18} & \good{+0.130} \\
\midrule
\textbf{Overall} & \textbf{462} & \textbf{0.255} & \textbf{0.590} & \textbf{0.774} & \textbf{57.71} & \textbf{0.754} \\
\bottomrule
\end{tabular}
}
\caption{Overall absolute scores, with per-script deltas shown in parentheses (row minus overall). Green indicates improvement (for CER: lower is better), red indicates degradation.}
\label{tab:results_script_en_target_deltas}
\end{table}

\paragraph{Effect of source script}
We next examine GPT-5.6 Luna on language pairs with English as the target language, grouped by the source writing system. Table~\ref{tab:results_script_en_target_deltas} reports per-script \emph{deltas} relative to the overall mean. Latin (the largest group) improves over the mean across all metrics. Cyrillic shows consistent gains in translation quality despite slightly worse OCR. Cyrillic and Latin patterns are also observed for Georgian and Chinese, respectively, albeit with much smaller $N$.

In contrast, Dravidian and Bengali-Assamese degrade across most metrics, indicating that these scripts remain challenging for both transcription and translation. Arabic further illustrates that the metrics do not necessarily move together: it exhibits better CER but worse localization and downstream translation, so accurate transcription alone is not sufficient. Overall, OCR accuracy, localization accuracy, and translation quality are only weakly coupled across scripts; results for very small groups should be interpreted cautiously.

\paragraph{Effects of visual modality on translation quality}
\label{sec:no_visual_modality}

%\begin{table}
%     \centering
%     \small
%     \resizebox{\columnwidth}{!}{
%     \begin{tabular}{lccc}
%     \toprule
%     Model & Img & No img & Text-only img \\
%     \midrule
%     claude-haiku-4.5 & 57.94 \scriptsize{[55.66, 60.16]} & 58.04 \scriptsize{[55.70, 60.28], $p$=0.417} & 57.60 \scriptsize{[55.29, 59.89], $p$=0.143} \\
%     gpt-5.6-luna & 62.30 \scriptsize{[59.99, 64.69]} & 62.29 \scriptsize{[60.15, 64.57], $p$=0.497} & 62.46 \scriptsize{[60.18, 64.79], $p$=0.309} \\
%     kimi-k3 & 62.95 \scriptsize{[60.58, 65.37]} & 62.76 \scriptsize{[60.45, 64.96], $p$=0.360} & 63.08 \scriptsize{[60.80, 65.41], $p$=0.491} \\
%     qwen3.5-122b-a10b & 57.43 \scriptsize{[55.14, 59.74]} & 60.18$^{*}$ \scriptsize{[57.97, 62.57], $p$<0.001} & 57.84 \scriptsize{[55.48, 60.31], $p$=0.268} \\
%     \bottomrule
%     \end{tabular}}
%     \caption{chrF scores on the test split, computed with unaligned (line-order) matching, with 95\% bootstrap confidence intervals and paired bootstrap significance testing against the baseline condition.
%     $^{*}$ means significant at $\alpha=0.05$. All of the conditions use the gold, extracted texts from the SVGs, \emph{Image} additionally uses the source image, \emph{Text-only Image} additionally uses only the text layer of the input image,  and \emph{No-Image} translates the extracted text without ever seeing the image.}
%\label{tab:chrf-scores}
%     \end{table}

\begin{table}
     \centering
     \small
     \resizebox{\columnwidth}{!}{
     \begin{tabular}{lccc}
     \toprule
     Model & Image & No image & Text-only image \\
     \midrule
     Claude Haiku 4.5 & 57.94 & 58.04 & 57.60 \\
     GPT-5.6 Luna & 62.30 & 62.29 & 62.46 \\
     Kimi K3 & 62.95 & 62.76 & 63.08 \\
     Qwen3.5 & 57.43 & 60.18$^{*}$ & 57.84 \\
     \bottomrule
     \end{tabular}}
     \caption{chrF scores on the test split, computed with unaligned (line-order) matching, with paired bootstrap significance testing against the baseline condition.
     $^{*}$ means significant at $\alpha=0.05$. All of the conditions use the gold, extracted texts from the SVGs, \emph{Image} additionally uses the source image, \emph{Text-only image} additionally uses only the text layer of the input image,  and \emph{No image} translates the extracted text without ever seeing the image.}
\label{tab:chrf-scores-no-ci}
     \end{table}

\Cref{tab:chrf-scores-no-ci} reports chrF scores under three input conditions: \emph{Image} (full image), \emph{Text-only image} (text layer only), and \emph{No image}.
As apparent from the table, differences across conditions are small. Overall, providing the image does not materially change performance: either the extracted texts are already unambiguous, or these models are not reliably using the visual signal for disambiguation.

%\formn{Keep or lose the following paragraphs? My opinion would be to lose them, the score differences are probably not significant and discussing them is not useful, Pepa.}
If we nevertheless treat the small differences as meaningful, the \emph{Image} condition is never best for any of the four models and is worst (or tied-worst) for Claude Haiku 4.5, Kimi K3, and Qwen3.5. The strongest condition depends on the model: \emph{No image} is best for Claude Haiku 4.5 and Qwen3.5, while \emph{Text-only image} is best for GPT-5.6 Luna and Kimi K3.

For Claude Haiku 4.5, GPT-5.6 Luna, and Kimi K3, the spread between the best and worst condition is at most 0.49 chrF, plausibly within run-to-run noise.

Overall, these results run counter to our hypothesis in \cref{sec:effect_of_image}: conditioning on the source image is neutral-to-harmful for chrF across all four models. This may reflect competition for attention/context budget or OCR/reading errors from visual clutter; since infographic text is typically unambiguous, images offer little disambiguation benefit. We conclude that adding the source image is not a reliable way to improve translation quality on this benchmark.

\section{Conclusion}
\label{sec:concl}

In this paper, we introduced \textsc{ImgMT}, a massively multilingual dataset for evaluating text-in-image translation (image-only inputs), such as Text Image Translation (text-only output).
We also proposed a multi-level evaluation protocol that combines complementary metrics at different granularities.
We demonstrated the framework on several MLLMs and one dedicated system in both English$\to$X and X$\to$English settings, showing that the source language and writing system matter for TIT quality.
Finally, we used \textsc{ImgMT} to probe the role of visual context in multimodal translation (text+image input). In our setup, adding the source image yields little benefit over text-only inputs and can even degrade performance, suggesting that \textsc{ImgMT} is not well suited for isolating the contribution of visual information in MMT.

Our experiments focused mainly on TIT; however, \textsc{ImgMT} can also be used to benchmark In-image Machine Translation (image-only output), which we leave for future work.

\section*{Limitations}

\paragraph{Data and evaluation.}
The dataset is derived from Wikimedia Commons SVGs, which introduces limitations for both data quality and evaluation. First, the annotations do not capture all text present in the source images. To reduce noise, we extract only text represented as SVG text spans and do not apply OCR; as a result, text rendered as vector paths is missing. Moreover, we rely on Inkscape's spatial query tool, which requires text elements to have explicit \texttt{id} attributes; elements without an \texttt{id} (or embedded in complex paths/groups) cannot be queried for exact bounding boxes and are omitted to avoid alignment errors. These choices may bias evaluation (e.g., by overestimating recall). Second, the language-pair distribution is imbalanced: because the source images come from Wikimedia Commons, high-resource languages (e.g., English, Spanish, French, German, and Russian) dominate, while some low-resource or regional languages occur in fewer than 10 pairs. Finally, because \textsc{ImgMT} is built from publicly available Wikimedia data, there is a risk of data contamination: some tested MLLMs may have encountered parts of the benchmark during pre-training.

\paragraph{Experiments.}
Due to budget and computational constraints, we restricted our evaluation to translation directions that include English (English$\to$X and X$\to$English) in a zero-shot setting, and we evaluated only the test split. This means our results cover only a subset of the available benchmark pairs; the remaining roughly 30,000 pairs in the training and development splits were prepared primarily to enable future training or fine-tuning studies, but we did not evaluate all bidirectional pairs across the full dataset in this paper. In addition, while \textsc{ImgMT} is suitable for evaluating text-in-image translation approaches beyond TIT, including IIMT (image-only output), we did not benchmark the IIMT setup here.

\section*{Acknowledgments}
% Sarka, cestovne:
This work was supported by the project TQ01000458 (EdUKate) financed by the Technology Agency of the Czech Republic (www.tacr.cz) within the Sigma 3 Programme.
It has been also partially supported by
the grant TQ28000003 (NCUI, Národní centrum umělé inteligence),   % Ondrej 
and
% Mezisektor: Josef
the grants CZ.02.01.01/00/23\_020/0008518 (“Jazykověda, umělá inteligence a jazykové a řečové technologie: od výzkumu k aplikacím”) 
% Michal: LINDAT/CLARIAH-CZ
and LM2023062 LINDAT/CLARIAH-CZ,
both financed by the Ministry of Education, Youth and Sports of the Czech Republic.

We also thank the three anonymous reviewers for their helpful comments and suggestions.

%\bibliography{custom,latex/anthology-1,latex/anthology-2}
\bibliography{main}

\begin{thebibliography}{27}
\providecommand{\natexlab}[1]{#1}

\bibitem[{Caglayan et~al.(2019)Caglayan, Madhyastha, Specia, and Barrault}]{caglayan-etal-2019-probing}
Ozan Caglayan, Pranava Madhyastha, Lucia Specia, and Lo{\"i}c Barrault. 2019.
\newblock \href {https://doi.org/10.18653/v1/N19-1422} {Probing the need for visual context in multimodal machine translation}.
\newblock In \emph{Proceedings of the 2019 Conference of the North {A}merican Chapter of the Association for Computational Linguistics: Human Language Technologies, Volume 1 (Long and Short Papers)}, pages 4159--4170, Minneapolis, Minnesota. Association for Computational Linguistics.

\bibitem[{Cui et~al.(2025)Cui, Sun, Lin, Gao, Zhang, Liu, Wang, Zhang, Zhou, Liu, Zhang, Lv, Huang, Zhang, Zhang, Zhang, Liu, Yu, and Ma}]{paddleocr}
Cheng Cui, Ting Sun, Manhui Lin, Tingquan Gao, Yubo Zhang, Jiaxuan Liu, Xueqing Wang, Zelun Zhang, Changda Zhou, Hongen Liu, Yue Zhang, Wenyu Lv, Kui Huang, Yichao Zhang, Jing Zhang, Jun Zhang, Yi~Liu, Dianhai Yu, and Yanjun Ma. 2025.
\newblock \href {https://arxiv.org/abs/2507.05595} {Paddleocr 3.0 technical report}.
\newblock \emph{Preprint}, arXiv:2507.05595.

\bibitem[{Elliott(2018)}]{elliott-2018-adversarial}
Desmond Elliott. 2018.
\newblock \href {https://doi.org/10.18653/v1/D18-1329} {Adversarial evaluation of multimodal machine translation}.
\newblock In \emph{Proceedings of the 2018 Conference on Empirical Methods in Natural Language Processing}, pages 2974--2978, Brussels, Belgium. Association for Computational Linguistics.

\bibitem[{Elliott et~al.(2016)Elliott, Frank, Sima{'}an, and Specia}]{elliott-etal-2016-multi30k}
Desmond Elliott, Stella Frank, Khalil Sima{'}an, and Lucia Specia. 2016.
\newblock \href {https://doi.org/10.18653/v1/W16-3210} {{M}ulti30{K}: Multilingual {E}nglish-{G}erman image descriptions}.
\newblock In \emph{Proceedings of the 5th Workshop on Vision and Language}, pages 70--74, Berlin, Germany. Association for Computational Linguistics.

\bibitem[{Futeral et~al.(2023)Futeral, Schmid, Laptev, Sagot, and Bawden}]{futeral.etal23}
Matthieu Futeral, Cordelia Schmid, Ivan Laptev, Beno{\^i}t Sagot, and Rachel Bawden. 2023.
\newblock \href {https://doi.org/10.18653/v1/2023.acl-long.295} {Tackling {{Ambiguity}} with {{Images}}: {{Improved Multimodal Machine Translation}} and {{Contrastive Evaluation}}}.
\newblock In \emph{Proceedings of the 61st {{Annual Meeting}} of the {{Association}} for {{Computational Linguistics}} ({{Volume}} 1: {{Long Papers}})}, pages 5394--5413, Toronto, Canada. Association for Computational Linguistics.

\bibitem[{Guerra and Khrapunova(2026)}]{guerra-khrapunova-2026-picture}
Vera~Senderowicz Guerra and Olesia Khrapunova. 2026.
\newblock \href {https://aclanthology.org/2026.eamt-2.29/} {Is a picture worth a thousand words? exploration and implementation considerations for visual context in translation workflows}.
\newblock In \emph{Proceedings of the 26th Annual Conference of the {E}uropean Association for Machine Translation (Volume 2)}, pages 91--103, Tilburg, The Netherlands. European Association for Machine Translation.

\bibitem[{{Inkscape Project}(2023)}]{inkscape}
{Inkscape Project}. 2023.
\newblock \href {https://inkscape.org/} {Inkscape}.
\newblock Version 1.3.2.

\bibitem[{Lan et~al.(2024)Lan, Niu, Meng, Zhou, Zhang, and Su}]{lan.etal24}
Zhibin Lan, Liqiang Niu, Fandong Meng, Jie Zhou, Min Zhang, and Jinsong Su. 2024.
\newblock \href {https://doi.org/10.18653/v1/2024.findings-acl.325} {Translatotron-{{V}}(ison): {{An End-to-End Model}} for {{In-Image Machine Translation}}}.
\newblock In \emph{Findings of the {{Association}} for {{Computational Linguistics}}: {{ACL}} 2024}, pages 5472--5485, Bangkok, Thailand. Association for Computational Linguistics.

\bibitem[{Lan et~al.(2023)Lan, Yu, Li, Zhang, Luan, Wang, Huang, and Su}]{lan.etal23}
Zhibin Lan, Jiawei Yu, Xiang Li, Wen Zhang, Jian Luan, Bin Wang, Degen Huang, and Jinsong Su. 2023.
\newblock \href {https://doi.org/10.18653/v1/2023.acl-long.192} {Exploring {{Better Text Image Translation}} with {{Multimodal Codebook}}}.
\newblock In \emph{Proceedings of the 61st {{Annual Meeting}} of the {{Association}} for {{Computational Linguistics}} ({{Volume}} 1: {{Long Papers}})}, pages 3479--3491, Toronto, Canada. Association for Computational Linguistics.

\bibitem[{Li et~al.(2025)Li, Zhu, and Wen}]{li.etal25}
Bo~Li, Shaolin Zhu, and Lijie Wen. 2025.
\newblock {{MIT-10M}}: {{A Large Scale Parallel Corpus}} of {{Multilingual Image Translation}}.
\newblock In \emph{Proceedings of the 31st {{International Conference}} on {{Computational Linguistics}}}, pages 5154--5167, Abu Dhabi, UAE. Association for Computational Linguistics.

\bibitem[{Li et~al.(2026)Li, Zhang, Liang, Shen, Zhang, Lyu, Wang, Wan, Zeng, Hu, Ma, and Zhou}]{Li_2026_CVPR}
Gengluo Li, Chengquan Zhang, Yupu Liang, Huawen Shen, Yaping Zhang, Pengyuan Lyu, Weinong Wang, Xingyu Wan, Gangyan Zeng, Han Hu, Can Ma, and Yu~Zhou. 2026.
\newblock {{MMTIT-bench}}: A multilingual and multi-scenario benchmark with cognition-perception-reasoning guided text-image machine translation.
\newblock In \emph{Proceedings of the {{IEEE}}/{{CVF}} Conference on Computer Vision and Pattern Recognition ({{CVPR}})}, pages 16593--16602.

\bibitem[{Liang et~al.(2024)Liang, Zhang, Ma, Zhang, Zhao, Xiang, Zong, and Zhou}]{liang.etal24}
Yupu Liang, Yaping Zhang, Cong Ma, Zhiyang Zhang, Yang Zhao, Lu~Xiang, Chengqing Zong, and Yu~Zhou. 2024.
\newblock \href {https://doi.org/10.18653/v1/2024.naacl-long.392} {Document {{Image Machine Translation}} with {{Dynamic Multi-pre-trained Models Assembling}}}.
\newblock In \emph{Proceedings of the 2024 {{Conference}} of the {{North American Chapter}} of the {{Association}} for {{Computational Linguistics}}: {{Human Language Technologies}} ({{Volume}} 1: {{Long Papers}})}, pages 7084--7095, Mexico City, Mexico. Association for Computational Linguistics.

\bibitem[{Liu et~al.(2025)Liu, Kong, Sun, Patil, Vajpayee, Liu, Bhat, and Sadoughi}]{liu-etal-2025-detect}
Danyang Liu, Fanjie Kong, Xiaohang Sun, Dhruva Patil, Avijit Vajpayee, Zhu Liu, Vimal Bhat, and Najmeh Sadoughi. 2025.
\newblock \href {https://doi.org/10.18653/v1/2025.naacl-long.74} {Detect, disambiguate, and translate: On-demand visual reasoning for multimodal machine translation with large vision-language models}.
\newblock In \emph{Proceedings of the 2025 Conference of the Nations of the Americas Chapter of the Association for Computational Linguistics: Human Language Technologies (Volume 1: Long Papers)}, pages 1559--1570, Albuquerque, New Mexico. Association for Computational Linguistics.

\bibitem[{Lyu et~al.(2025)Lyu, Fu, Li, Zeng, Yang, Ma, Zhou, Luo, and Luan}]{lyu.etal25}
Jiahao Lyu, Pei Fu, Zhenhang Li, Weichao Zeng, Jiahui Yang, Can Ma, Yu~Zhou, Zhenbo Luo, and Jian Luan. 2025.
\newblock Immersive {{Multimodal Translation}}: {{A Proxy Task}} for {{Cross-modal}} and {{Objective Evaluation}} of {{Unified Models}}.

\bibitem[{Ma et~al.(2022)Ma, Zhang, Tu, Han, Wu, Zhao, and Zhou}]{ma.etal22}
Cong Ma, Yaping Zhang, Mei Tu, Xu~Han, Linghui Wu, Yang Zhao, and Yu~Zhou. 2022.
\newblock \href {https://doi.org/10.1109/ICPR56361.2022.9956695} {Improving {{End-to-End Text Image Translation From}} the {{Auxiliary Text Translation Task}}}.
\newblock In \emph{2022 26th {{International Conference}} on {{Pattern Recognition}} ({{ICPR}})}, pages 1664--1670.

\bibitem[{Parida et~al.(2024)Parida, Bojar, Abdulmumin, Muhammad, and Ahmad}]{parida-etal-2024-findings}
Shantipriya Parida, Ond{\v{r}}ej Bojar, Idris Abdulmumin, Shamsuddeen~Hassan Muhammad, and Ibrahim~Said Ahmad. 2024.
\newblock \href {https://doi.org/10.18653/v1/2024.wmt-1.56} {Findings of {WMT}2024 {E}nglish-to-low resource multimodal translation task}.
\newblock In \emph{Proceedings of the Ninth Conference on Machine Translation}, pages 677--683, Miami, Florida, USA. Association for Computational Linguistics.

\bibitem[{Popovi{\'c}(2015)}]{popovic2015chrf}
Maja Popovi{\'c}. 2015.
\newblock chrf: character n-gram f-score for automatic mt evaluation.
\newblock In \emph{Proceedings of the tenth workshop on statistical machine translation}, pages 392--395.

\bibitem[{Qian et~al.(2024)Qian, Zhang, Yang, Fan, Ma, Wong, Sun, and Ji}]{qian.etal24}
Zhipeng Qian, Pei Zhang, Baosong Yang, Kai Fan, Yiwei Ma, Derek~F. Wong, Xiaoshuai Sun, and Rongrong Ji. 2024.
\newblock \href {https://doi.org/10.18653/v1/2024.findings-emnlp.137} {{{AnyTrans}}: {{Translate AnyText}} in the {{Image}} with {{Large Scale Models}}}.
\newblock In \emph{Findings of the {{Association}} for {{Computational Linguistics}}: {{EMNLP}} 2024}, pages 2432--2444, Miami, Florida, USA. Association for Computational Linguistics.

\bibitem[{Rei et~al.(2022)Rei, C.~de Souza, Alves, Zerva, Farinha, Glushkova, Lavie, Coheur, and Martins}]{rei-etal-2022-comet}
Ricardo Rei, Jos{\'e}~G. C.~de Souza, Duarte Alves, Chrysoula Zerva, Ana~C Farinha, Taisiya Glushkova, Alon Lavie, Luisa Coheur, and Andr{\'e} F.~T. Martins. 2022.
\newblock \href {https://doi.org/10.18653/v1/2022.wmt-1.52} {{COMET}-22: Unbabel-{IST} 2022 submission for the metrics shared task}.
\newblock In \emph{Proceedings of the Seventh Conference on Machine Translation (WMT)}, pages 578--585, Abu Dhabi, United Arab Emirates (Hybrid). Association for Computational Linguistics.

\bibitem[{Srinivasan et~al.(2021)Srinivasan, Raman, Chen, Bendersky, and Najork}]{srinivasan.etal21}
Krishna Srinivasan, Karthik Raman, Jiecao Chen, Michael Bendersky, and Marc Najork. 2021.
\newblock \href {https://doi.org/10.1145/3404835.3463257} {{{WIT}}: {{Wikipedia-based Image Text Dataset}} for {{Multimodal Multilingual Machine Learning}}}.
\newblock In \emph{Proceedings of the 44th {{International ACM SIGIR Conference}} on {{Research}} and {{Development}} in {{Information Retrieval}}}, {{SIGIR}} '21, pages 2443--2449, New York, NY, USA. Association for Computing Machinery.

\bibitem[{Tian et~al.(2025)Tian, Liu, Liu, and Guo}]{tian.etal25}
Yanzhi Tian, Zeming Liu, Zhengyang Liu, and Yuhang Guo. 2025.
\newblock \href {https://doi.org/10.18653/v1/2025.findings-acl.6} {Exploring {{In-Image Machine Translation}} with {{Real-World Background}}}.
\newblock In \emph{Findings of the {{Association}} for {{Computational Linguistics}}: {{ACL}} 2025}, pages 124--137, Vienna, Austria. Association for Computational Linguistics.

\bibitem[{Vaidya et~al.(2025)Vaidya, Sharma, Gatti, and Mishra}]{vaidya.etal25}
Shreyas Vaidya, Arvind~Kumar Sharma, Prajwal Gatti, and Anand Mishra. 2025.
\newblock \href {https://doi.org/10.1007/978-3-031-78113-1_21} {Show {{Me}} the {{World}} in {{My Language}}: {{Establishing}} the {{First Baseline}} for {{Scene-Text}} to {{Scene-Text Translation}}}.
\newblock In \emph{Pattern {{Recognition}}}, pages 312--328, Cham. Springer Nature Switzerland.

\bibitem[{Wang et~al.(2004)Wang, Bovik, Sheikh, and Simoncelli}]{wang2004ssim}
Zhou Wang, A.C. Bovik, H.R. Sheikh, and E.P. Simoncelli. 2004.
\newblock \href {https://doi.org/10.1109/TIP.2003.819861} {Image quality assessment: from error visibility to structural similarity}.
\newblock \emph{IEEE Transactions on Image Processing}, 13(4):600--612.

\bibitem[{Wu et~al.(2026)Wu, Han, Li, Xu, Zhao, Jin, Wang, and Luo}]{wu.etal26}
Qingyu Wu, Yuxuan Han, Haijun Li, Zhao Xu, Jianshan Zhao, Xu~Jin, Longyue Wang, and Weihua Luo. 2026.
\newblock \href {https://doi.org/10.48550/arXiv.2602.07014} {Vectra: {{A New Metric}}, {{Dataset}}, and {{Model}} for {{Visual Quality Assessment}} in {{E-Commerce In-Image Machine Translation}}}.
\newblock \emph{Preprint}, arXiv:2602.07014.

\bibitem[{Wu et~al.(2021)Wu, Kong, Bi, Li, and Kao}]{wu-etal-2021-good}
Zhiyong Wu, Lingpeng Kong, Wei Bi, Xiang Li, and Ben Kao. 2021.
\newblock \href {https://doi.org/10.18653/v1/2021.acl-long.480} {Good for misconceived reasons: An empirical revisiting on the need for visual context in multimodal machine translation}.
\newblock In \emph{Proceedings of the 59th Annual Meeting of the Association for Computational Linguistics and the 11th International Joint Conference on Natural Language Processing (Volume 1: Long Papers)}, pages 6153--6166, Online. Association for Computational Linguistics.

\bibitem[{Yu et~al.(2025)Yu, Sun, Zhao, Song, and Yang}]{yu-etal-2025-imagination}
Zhuang Yu, Shiliang Sun, Jing Zhao, Tengfei Song, and Hao Yang. 2025.
\newblock \href {https://doi.org/10.18653/v1/2025.findings-emnlp.579} {Imagination and contemplation: A balanced framework for semantic-augmented multimodal machine translation}.
\newblock In \emph{Findings of the Association for Computational Linguistics: EMNLP 2025}, pages 10913--10928, Suzhou, China. Association for Computational Linguistics.

\bibitem[{Zhu et~al.(2023)Zhu, Li, Lei, and Xiong}]{zhu.etal23}
Shaolin Zhu, Shangjie Li, Yikun Lei, and Deyi Xiong. 2023.
\newblock \href {https://doi.org/10.18653/v1/2023.acl-long.751} {{{PEIT}}: {{Bridging}} the {{Modality Gap}} with {{Pre-trained Models}} for {{End-to-End Image Translation}}}.
\newblock In \emph{Proceedings of the 61st {{Annual Meeting}} of the {{Association}} for {{Computational Linguistics}} ({{Volume}} 1: {{Long Papers}})}, pages 13433--13447, Toronto, Canada. Association for Computational Linguistics.

\end{thebibliography}

\appendix
\clearpage

\onecolumn

\section{Dataset Structure}
\label{sec:app-dataset-structure}

\begin{center}
    \begin{tcolorbox}[
        colframe=black,
        colback=white,
        boxrule=0.15mm,
        width=0.98\linewidth,
        arc=0mm,
        auto outer arc,
        boxsep=1pt,
        left=2pt,right=2pt,top=2pt,bottom=2pt
    ]
    \scriptsize
\begin{verbatim}
data/
|-- train/
|   |-- {languageGroupNumber}/
|   |   |-- svg/
|   |   |   |-- {langCode}.svg
|   |   |   |-- ...
|   |   |-- png-base/
|   |   |   |-- {langCode}.png
|   |   |-- {sourceLangCode}-{targetLangCode}.json
|   |   |-- ...
|   |-- ...
|-- dev/
|   |-- Same structure as train
|-- test/
|   |-- Same structure as train
\end{verbatim}
    \end{tcolorbox}
    \captionof{figure}{Overall dataset directory layout}
    \label{fig:datasetDirectoryStructure}

    \vspace{0.3em}
    \begin{tcolorbox}[
        colframe=black,
        colback=white,
        boxrule=0.15mm,
        width=\linewidth,
        arc=0mm,
        auto outer arc,
        boxsep=1pt,
        left=2pt,right=2pt,top=2pt,bottom=2pt
    ]
    \scriptsize
    \begin{lstlisting}[
        basicstyle=\ttfamily\scriptsize,
        columns=fullflexible,
        keepspaces=true,
        showstringspaces=false,
        lineskip=-1.5pt,
        aboveskip=0pt,
        belowskip=0pt
    ]
{
    "alignment_method": string,
    "source_language": string,
    "source_SVG": {
        "size": {
            "width": px,
            "height": px
        },
        "wikimedia_url": url,
        "licensing": {
            "license": string,
            "original_license": string,
            "author": string
        }
    },
    "source_texts": [ string ],
    "source_text_bounding_boxes": [
        {
            "x": float,
            "y": float,
            "w": float,
            "h": float
        }
    ],
    "target_language": language code,
    "target_SVG": {
        "size": {
            "width": px,
            "height": px
        },
        "wikimedia_url": url,
        "licensing": {
            "license": string,
            "original_license": string,
            "author": string
        }
    }
    "target_texts": [ string ],
    "target_text_bounding_boxes": [
        {
            "x": float,
            "y": float,
            "w": float,
            "h": float
        }
    ],
}
\end{lstlisting}
\end{tcolorbox}
    \captionof{figure}{JSON structure}
    \label{fig:datasetStructure}
\end{center}

\clearpage
\section{Language and Script Distribution}
\label{app:langs}

\begin{longtable}{ll r r p{5.5cm}}
\toprule
\textbf{Lang} & \textbf{Writing System} & \textbf{Pairs} & \textbf{Targets} & \textbf{Top Paired Languages} \\
\midrule
\endfirsthead
\multicolumn{5}{c}%
{{\bfseries \tablename\ \thetable{} -- continued from previous page}} \\
\toprule
\textbf{Lang} & \textbf{Writing System} & \textbf{Pairs} & \textbf{Targets} & \textbf{Top Paired Languages} \\
\midrule
\endhead
\midrule
\multicolumn{5}{r}{{Continued on next page}} \\
\bottomrule
\endfoot
\bottomrule
\endlastfoot
\texttt{en} & Latin & 5,089 & 121 & \small es (386), fr (385), ru (325), de (271), it (187) \\
\texttt{es} & Latin & 3,461 & 122 & \small en (386), fr (274), pt (191), ru (177), de (171) \\
\texttt{fr} & Latin & 3,253 & 124 & \small en (385), es (274), de (184), ru (166), it (150) \\
\texttt{ru} & Cyrillic & 2,591 & 117 & \small en (325), es (177), fr (166), de (139), it (106) \\
\texttt{de} & Latin & 2,568 & 121 & \small en (271), fr (184), es (171), ru (139), it (95) \\
\texttt{it} & Latin & 2,229 & 114 & \small en (187), es (169), fr (150), ru (106), de (95) \\
\texttt{ca} & Latin & 1,960 & 112 & \small en (171), es (155), fr (104), it (76), de (75) \\
\texttt{hi} & Devanagari & 1,728 & 109 & \small en (141), te (132), ml (103), gu (100), kn (87) \\
\texttt{hu} & Latin & 1,707 & 108 & \small en (163), es (139), fr (109), de (84), ru (82) \\
\texttt{pt} & Latin & 1,705 & 111 & \small es (191), en (142), fr (101), ru (85), it (77) \\
\texttt{pl} & Latin & 1,590 & 111 & \small en (112), es (88), de (75), fr (73), ru (66) \\
\texttt{te} & Dravidian & 1,570 & 81 & \small en (147), ml (136), hi (132), gu (113), kn (98) \\
\texttt{ml} & Dravidian & 1,537 & 106 & \small te (136), en (125), hi (103), gu (97), kn (88) \\
\texttt{ta} & Dravidian & 1,345 & 102 & \small en (91), te (79), hi (68), ml (67), kn (60) \\
\texttt{gu} & Gujarati & 1,322 & 74 & \small te (113), en (112), hi (100), ml (97), as (75) \\
\texttt{cs} & Latin & 1,313 & 112 & \small en (87), fr (76), es (67), de (54), it (53) \\
\texttt{uk} & Cyrillic & 1,299 & 111 & \small en (92), ru (72), de (61), fr (61), es (60) \\
\texttt{as} & Bengali-Assamese & 1,292 & 86 & \small en (100), te (83), hi (82), ml (75), gu (75) \\
\texttt{bn} & Bengali-Assamese & 1,231 & 105 & \small en (106), hi (82), te (81), ml (72), gu (64) \\
\texttt{ar} & Arabic & 1,144 & 109 & \small en (101), es (78), fr (55), it (47), ru (45) \\
\texttt{tr} & Latin & 1,072 & 109 & \small en (77), fr (50), ru (50), es (48), de (38) \\
\texttt{pa} & Gurmukhi & 1,053 & 75 & \small te (89), en (83), hi (83), ml (77), kn (68) \\
\texttt{id} & Latin & 1,031 & 115 & \small en (71), es (49), fr (44), de (37), it (34) \\
\texttt{kn} & Dravidian & 1,025 & 68 & \small te (98), ml (88), hi (87), en (87), gu (74) \\
\texttt{nl} & Latin & 1,000 & 110 & \small en (72), es (65), fr (53), de (50), it (41) \\
\texttt{zh} & Chinese & 984 & 112 & \small en (80), es (68), de (53), fr (52), ru (37) \\
\texttt{mkd} & Cyrillic & 896 & 84 & \small en (118), fr (59), es (59), ru (46), de (39) \\
\texttt{sv} & Latin & 892 & 106 & \small en (52), es (42), fr (40), de (38), it (37) \\
\texttt{ja} & Japanese & 777 & 106 & \small en (46), fr (36), es (32), ru (29), de (28) \\
\texttt{ro} & Latin & 758 & 109 & \small en (36), es (34), fr (32), de (30), it (26) \\
\texttt{eo} & Latin & 750 & 104 & \small en (37), es (34), fr (32), de (29), ca (27) \\
\texttt{ko} & Korean & 730 & 110 & \small en (34), fr (34), es (33), ca (25), de (23) \\
\texttt{he} & Hebrew & 727 & 102 & \small en (46), es (38), fr (38), de (28), ru (28) \\
\texttt{fi} & Latin & 726 & 104 & \small en (47), es (34), fr (33), de (32), it (24) \\
\texttt{mr} & Devanagari & 720 & 73 & \small hi (59), gu (57), en (53), te (52), ml (46) \\
\texttt{gl} & Latin & 698 & 108 & \small en (39), es (32), it (29), fr (28), de (27) \\
\texttt{fa} & Arabic & 612 & 110 & \small en (23), es (21), uk (19), fr (19), cs (18) \\
\texttt{bg} & Cyrillic & 587 & 102 & \small en (32), es (26), fr (26), de (21), it (19) \\
\texttt{da} & Latin & 582 & 101 & \small en (28), es (24), fr (23), de (20), ru (19) \\
\texttt{ka} & Georgian & 565 & 103 & \small en (78), ru (27), es (25), hu (22), it (20) \\
\texttt{eu} & Latin & 565 & 104 & \small en (27), es (23), fr (23), ca (20), ru (19) \\
\texttt{lt} & Latin & 565 & 110 & \small en (22), ca (20), es (20), fr (20), ru (19) \\
\texttt{no} & Latin & 553 & 94 & \small en (36), es (29), fr (26), ru (25), hu (23) \\
\texttt{hr} & Latin & 545 & 98 & \small en (41), es (27), fr (26), hu (20), de (19) \\
\texttt{lv} & Latin & 538 & 99 & \small de (24), fr (22), en (20), es (19), cs (18) \\
\texttt{sat} & Ol\_Chiki & 525 & 55 & \small te (45), en (45), gu (41), hi (41), ml (41) \\
\texttt{vi} & Latin & 517 & 106 & \small en (30), es (25), fr (25), de (20), ca (17) \\
\texttt{sl} & Latin & 454 & 100 & \small en (28), es (16), ru (15), de (15), fr (15) \\
\texttt{sk} & Latin & 439 & 103 & \small fr (21), en (20), de (17), pl (15), cs (15) \\
\texttt{af} & Latin & 411 & 105 & \small en (21), fr (18), es (17), ca (14), de (14) \\
\texttt{be} & Cyrillic & 401 & 104 & \small en (21), fr (18), es (15), de (15), ru (14) \\
\texttt{ne} & Devanagari & 401 & 53 & \small en (29), gu (29), te (28), hi (27), bn (25) \\
\texttt{la} & Latin & 348 & 100 & \small en (17), fr (17), es (15), de (13), ru (12) \\
\texttt{zh\_hans} & Chinese & 340 & 66 & \small en (27), es (26), fr (17), it (17), de (16) \\
\texttt{bs} & Latin & 333 & 93 & \small cs (11), en (10), it (10), pl (9), ca (9) \\
\texttt{el} & Greek & 310 & 101 & \small fr (13), de (13), es (12), en (12), ca (11) \\
\texttt{az} & Latin & 287 & 93 & \small en (12), it (10), es (9), ru (9), pt (8) \\
\texttt{hy} & Armenian & 285 & 91 & \small en (18), fr (13), ru (12), es (12), de (10) \\
\texttt{et} & Latin & 285 & 91 & \small en (12), it (9), ru (9), pl (9), fr (9) \\
\texttt{nb} & Latin & 276 & 70 & \small fr (17), en (17), es (16), de (13), ca (12) \\
\texttt{zh\_hant} & Chinese & 273 & 71 & \small en (14), zh\_hans (13), it (11), es (11), fr (9) \\
\texttt{ur} & Arabic & 259 & 97 & \small en (12), fr (9), es (9), it (8), ar (8) \\
\texttt{is} & Latin & 250 & 94 & \small en (8), es (7), ru (7), fr (6), pl (6) \\
\texttt{sr} & Cyrillic & 249 & 93 & \small en (12), es (10), fr (10), de (9), ru (9) \\
\texttt{th} & Thai & 238 & 97 & \small fr (9), de (7), en (7), ca (7), ru (7) \\
\texttt{mk} & Cyrillic & 238 & 95 & \small de (8), fr (8), ca (8), en (7), ru (7) \\
\texttt{ast} & Latin & 214 & 90 & \small es (9), fr (8), hu (7), ca (7), be (6) \\
\texttt{dty} & Devanagari & 213 & 46 & \small ne (17), gu (14), as (13), hi (12), te (11) \\
\texttt{ku} & Latin & 205 & 70 & \small es (13), en (12), ca (11), it (8), de (8) \\
\texttt{my} & Burmese & 199 & 77 & \small id (7), en (7), de (6), fr (6), cs (6) \\
\texttt{lb} & Latin & 190 & 87 & \small fr (9), en (8), de (8), uk (7), pl (6) \\
\texttt{uz} & Cyrillic & 185 & 92 & \small pl (6), de (5), tr (5), es (5), fr (5) \\
\texttt{ga} & Latin & 184 & 86 & \small es (7), pl (5), hu (5), pt (4), sl (4) \\
\texttt{nn} & Latin & 182 & 83 & \small fr (7), nl (6), de (6), en (5), es (5) \\
\texttt{ms} & Latin & 175 & 88 & \small ca (13), it (8), es (6), ru (6), hi (5) \\
\texttt{an} & Latin & 172 & 84 & \small es (5), en (5), uk (4), it (4), pl (4) \\
\texttt{tl} & Latin & 164 & 94 & \small it (4), pt (4), es (4), cs (4), tr (4) \\
\texttt{br} & Latin & 163 & 67 & \small en (9), fr (8), de (7), es (6), ru (6) \\
\texttt{oc} & Latin & 146 & 85 & \small ca (7), de (6), it (5), fr (5), ru (5) \\
\texttt{or} & Dravidian & 119 & 41 & \small hi (10), gu (10), en (7), te (7), mr (6) \\
\texttt{sh} & Latin & 117 & 79 & \small fr (3), en (3), ro (2), th (2), hi (2) \\
\texttt{cy} & Latin & 116 & 42 & \small en (11), es (9), fr (9), hu (6), ca (5) \\
\texttt{xmf} & Georgian & 113 & 84 & \small fr (3), uz (3), ca (3), de (3), uk (2) \\
\texttt{wuu} & Chinese & 113 & 79 & \small sl (2), ka (2), fr (2), bg (2), pt (2) \\
\texttt{su} & Latin & 111 & 55 & \small en (6), fr (5), es (4), de (4), cs (4) \\
\texttt{sa} & Devanagari & 111 & 58 & \small en (7), hi (6), ca (4), ta (4), ml (4) \\
\texttt{dsb} & Latin & 110 & 82 & \small hsb (2), es (2), en (2), fr (2), be (2) \\
\texttt{hsb} & Latin & 110 & 82 & \small dsb (2), ko (2), it (2), de (2), hy (2) \\
\texttt{io} & Latin & 100 & 77 & \small uk (2), fr (2), eo (2), pt (2), zh (2) \\
\texttt{kk} & Cyrillic & 94 & 56 & \small en (4), it (4), cs (3), es (3), vi (3) \\
\texttt{smn} & Latin & 83 & 53 & \small en (6), fr (3), pt (3), de (3), it (3) \\
\texttt{ky} & Cyrillic & 81 & 77 & \small es (2), ca (2), lt (2), hu (2), de (1) \\
\texttt{sco} & Latin & 76 & 76 & \small hu (1), ar (1), hi (1), an (1), ro (1) \\
\texttt{rue} & Cyrillic & 76 & 76 & \small uk (1), sv (1), ko (1), az (1), eu (1) \\
\texttt{ext} & Latin & 76 & 76 & \small da (1), eo (1), hu (1), ast (1), ml (1) \\
\texttt{bxr} & Cyrillic & 76 & 76 & \small ms (1), da (1), an (1), vi (1), ml (1) \\
\texttt{gd} & Latin & 76 & 76 & \small nn (1), tt (1), wuu (1), ga (1), ru (1) \\
\texttt{tt} & Cyrillic & 76 & 76 & \small gd (1), ta (1), no (1), vi (1), cs (1) \\
\texttt{war} & Latin & 76 & 76 & \small sk (1), io (1), fr (1), vi (1), no (1) \\
\texttt{lmo} & Latin & 76 & 76 & \small sv (1), lv (1), bn (1), af (1), ka (1) \\
\texttt{si} & Sinhala & 71 & 55 & \small fr (3), en (3), de (3), es (3), bs (2) \\
\texttt{km} & Khmer & 67 & 47 & \small en (4), es (3), ru (3), gl (3), fr (3) \\
\texttt{ckb} & Arabic & 65 & 50 & \small es (3), zh (2), hi (2), he (2), it (2) \\
\texttt{ce} & Cyrillic & 63 & 32 & \small ru (8), es (6), en (6), uk (4), hu (3) \\
\texttt{mn} & Cyrillic & 62 & 51 & \small ru (2), be (2), pt (2), ca (2), bg (2) \\
\texttt{sw} & Latin & 57 & 41 & \small zh (3), it (3), en (3), de (3), ar (2) \\
\texttt{ba} & Cyrillic & 48 & 43 & \small fr (2), ko (2), de (2), es (2), en (2) \\
\texttt{yue} & Chinese & 43 & 31 & \small fr (4), pt (4), ml (2), en (2), ca (2) \\
\texttt{sd} & Arabic & 42 & 38 & \small fa (2), fr (2), ru (2), es (2), ce (1) \\
\texttt{fy} & Latin & 34 & 34 & \small fa (1), ckb (1), nl (1), pl (1), hi (1) \\
\texttt{nap} & Latin & 34 & 34 & \small sr (1), hsb (1), nl (1), gl (1), sk (1) \\
\texttt{ab} & Cyrillic & 25 & 25 & \small uz (1), ru (1), id (1), fi (1), ko (1) \\
\texttt{tg} & Cyrillic & 23 & 22 & \small tl (2), nb (1), zh (1), cs (1), shn (1) \\
\texttt{tok} & Latin & 21 & 21 & \small be (1), hu (1), no (1), ru (1), ro (1) \\
\texttt{ban} & Latin & 20 & 20 & \small de (1), la (1), uz (1), br (1), si (1) \\
\texttt{shn} & Shan & 20 & 20 & \small es (1), xmf (1), de (1), br (1), la (1) \\
\texttt{alb} & Latin & 19 & 12 & \small es (3), ru (2), en (2), fr (2), de (2) \\
\texttt{ks} & Arabic & 18 & 15 & \small zh (2), bn (2), hi (2), ar (1), id (1) \\
\texttt{pa\_guru} & Gurmukhi & 17 & 17 & \small gu (1), mr (1), ur (1), de (1), bn (1) \\
\texttt{ang} & Latin & 14 & 14 & \small en (1), tr (1), el (1), bjn (1), vi (1) \\
\texttt{bjn} & Arabic & 14 & 14 & \small mag (1), ang (1), id (1), tr (1), fr (1) \\
\texttt{mag} & Devanagari & 14 & 14 & \small bjn (1), ko (1), id (1), el (1), es (1) \\
\texttt{so} & Latin & 13 & 13 & \small de (1), af (1), sw (1), it (1), ca (1) \\
\texttt{arz} & Arabic & 13 & 12 & \small ja (2), be (1), pl (1), tr (1), fr (1) \\
\texttt{sr\_latn} & Latin & 6 & 6 & \small ca (1), eu (1), tr (1), ru (1), de (1) \\
\texttt{ps} & Arabic & 6 & 6 & \small ca (1), vi (1), fr (1), kk (1), fa (1) \\
\texttt{zh\_tw} & Chinese & 5 & 5 & \small id (1), fr (1), en (1), zh (1), ca (1) \\
\texttt{vec} & Latin & 4 & 4 & \small da (1), eu (1), sl (1), ro (1) \\
\texttt{se} & Latin & 4 & 4 & \small hi (1), ks (1), ms (1), fr (1) \\
\caption{Overview of all languages covered in the dataset, their writing systems, total translation pairs, number of unique paired target languages, and top paired counterpart languages.} \label{tab:all_languages} \\
\end{longtable}

\clearpage
\section{Tool for Manual Annotation}
\label{app:annottool}

\begin{figure*}[!h]
    \centering
    \includegraphics[width=\linewidth]{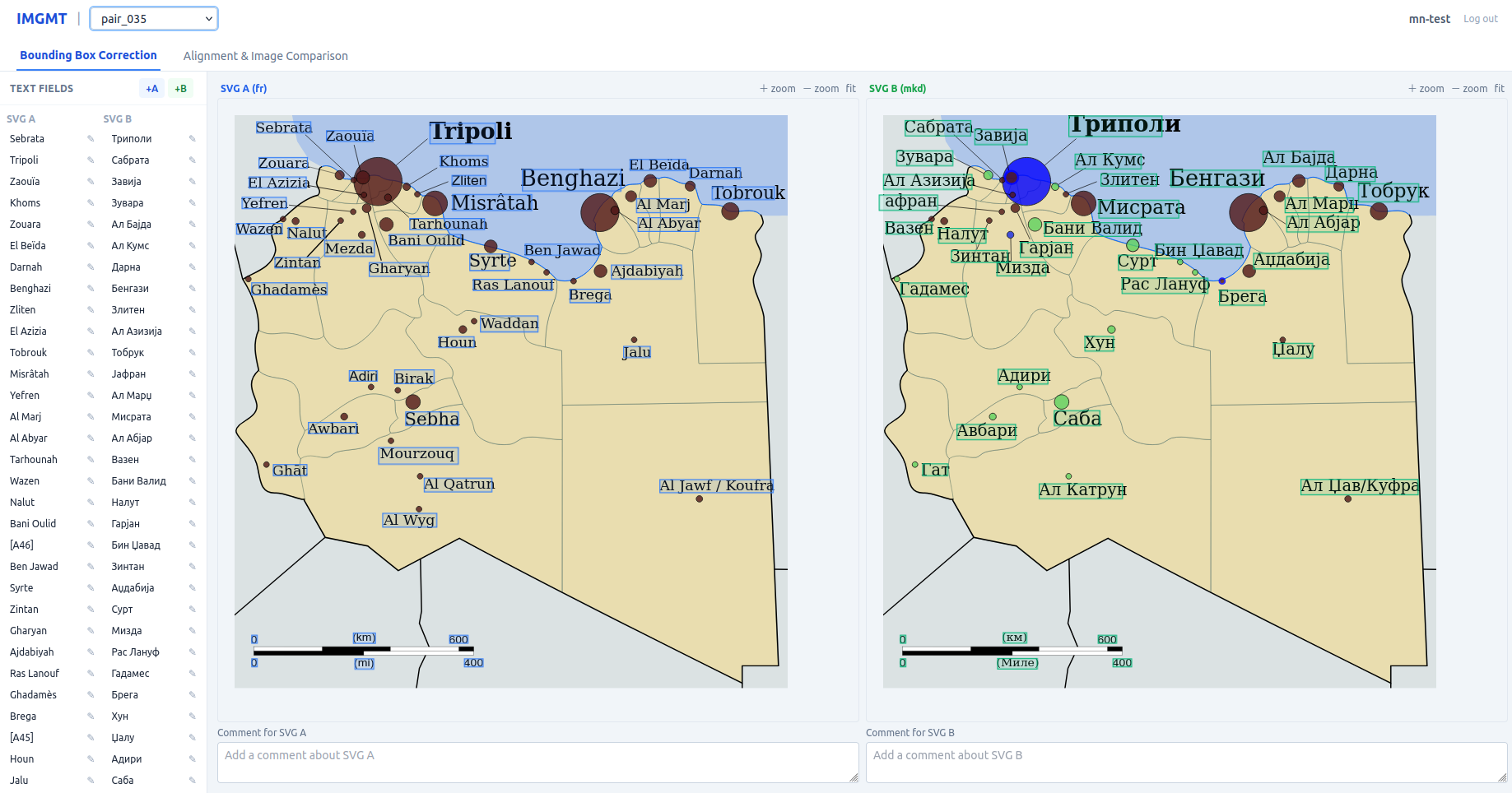}
    \caption{A screenshot of the tool used for manual annotation}
    \label{fig:placeholder}
\end{figure*}

\end{document}